\pdfoutput=1

\documentclass[11pt]{article}

\usepackage[preprint]{acl}

\usepackage{times}
\usepackage{latexsym}
\usepackage{booktabs}
\usepackage[most]{tcolorbox}
\usepackage{graphicx}
\newcommand{\ours}{\texttt{GLIMPSE}}

\usepackage[T1]{fontenc}

\usepackage[utf8]{inputenc}

\usepackage{microtype}

\usepackage{inconsolata}

\usepackage{graphicx}

\title{GLIMPSE: Do Large Vision-Language Models Truly \textit{Think With Videos} or Just Glimpse at Them?}

\author{
  \textbf{Yiyang Zhou*\textsuperscript{1}},
  \textbf{Linjie Li*\textsuperscript{2}},
  \textbf{Shi Qiu*\textsuperscript{1}},
  \textbf{Zhengyuan Yang\textsuperscript{2}},
\\
  \textbf{Yuyang Zhao\textsuperscript{3}},
  \textbf{Siwei Han\textsuperscript{1}},
  \textbf{Yangfan He\textsuperscript{4}},
  \textbf{Kangqi Li\textsuperscript{5}},
\\
  \textbf{Haonian Ji\textsuperscript{1}},
  \textbf{Zihao Zhao\textsuperscript{6}},
  \textbf{Haibo Tong\textsuperscript{7}},
  \textbf{Lijuan Wang\textsuperscript{2}},
  \textbf{Huaxiu Yao\textsuperscript{1}}
\\
  \textsuperscript{1}UNC Chapel-Hill,
  \textsuperscript{2}Microsoft, \textsuperscript{3}NUS, \textsuperscript{4}University of Minnesota\\ \textsuperscript{5}Case Western Reserve University
  \textsuperscript{6}Rutgers University, \textsuperscript{7}UIUC
\\
    \href{mailto:email@domain}{\{yiyangai, shiqiu, huaxiu\}@cs.unc.edu}
}

\begin{document}
\maketitle
\begin{abstract}

Existing video benchmarks often resemble image-based benchmarks, with question types like "What actions does the person perform throughout the video?" or "What color is the woman’s dress in the video?" For these, models can often answer by scanning just a few key frames, without deep temporal reasoning. This limits our ability to assess whether large vision-language models (LVLMs) can truly think with videos rather than perform superficial frame-level analysis. To address this, we introduce \ours, a benchmark specifically designed to evaluate whether LVLMs can genuinely think with videos. Unlike prior benchmarks, \ours\ emphasizes comprehensive video understanding beyond static image cues. It consists of 3,269 videos and over 4,342 highly visual-centric questions across 11 categories, including Trajectory Analysis, Temporal Reasoning, and Forensics Detection. All questions are carefully crafted by human annotators and require watching the entire video and reasoning over full video context—this is what we mean by thinking with video. These questions cannot be answered by scanning selected frames or relying on text alone. In human evaluations, \ours\ achieves 94.82\% accuracy, but current LVLMs face significant challenges. Even the best-performing model, GPT-o3, reaches only 66.43\%, highlighting that LVLMs still struggle to move beyond surface-level reasoning to truly think with videos. We publicly release our benchmark and code at \url{https://github.com/aiming-lab/GLIMPSE}.
\end{abstract}

\section{Introduction}
\label{sec:intro}
\vspace{-0.5em}

\begin{figure}
    \centering
    \includegraphics[width=0.95\linewidth]{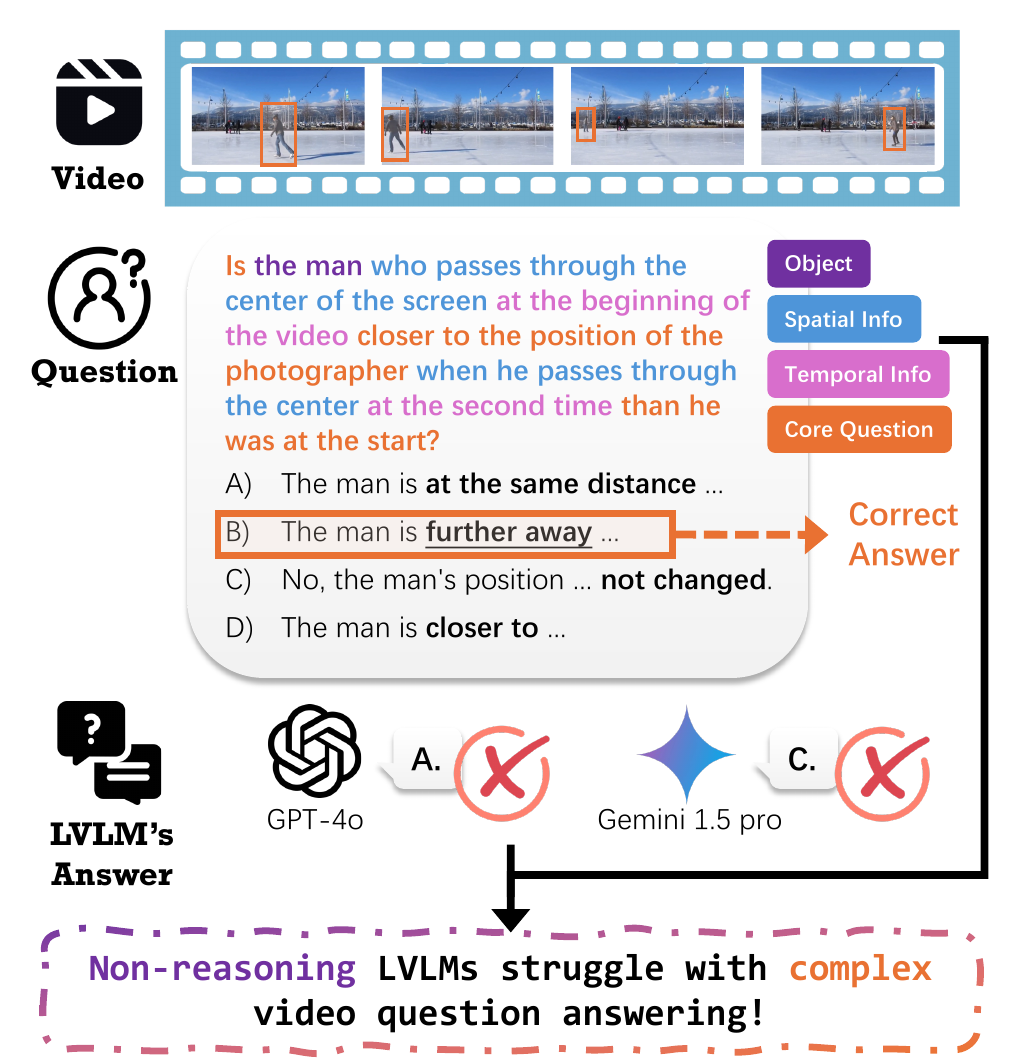}
    \caption{LVLMs without reasoning ability struggle with complex video question answering.}
    \label{fig:teaser}
\end{figure}
\begin{figure*}[t]
  \centering
  \includegraphics[width=0.95\textwidth]{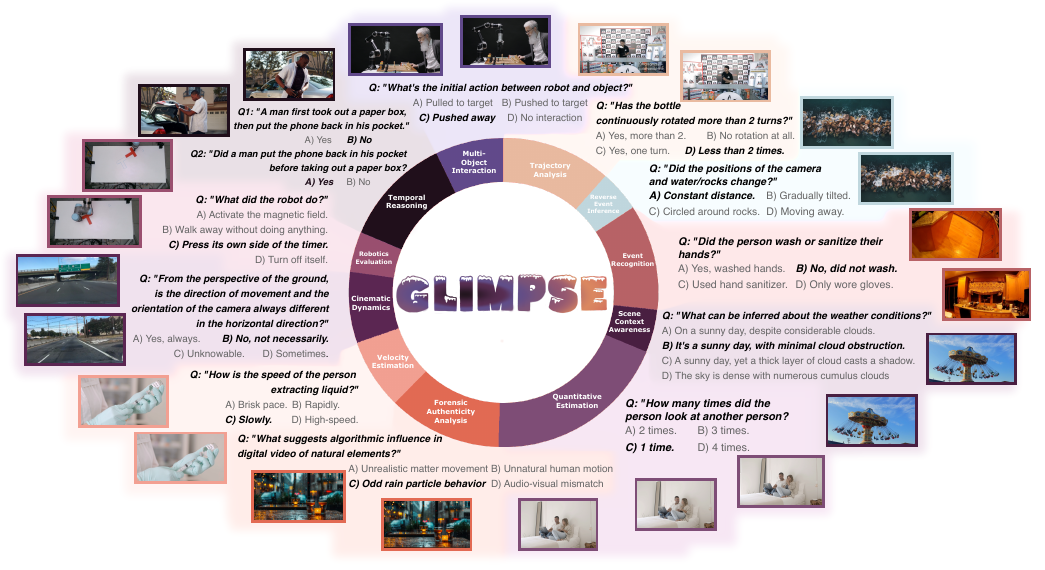}
  \vspace{-10pt}
\caption{\ours\ categorizes visual-centric questions in video data into 11 distinct types, with representative examples from each category illustrated in the figure. Human-annotated questions and answers are reformatted into multiple-choice format for LVLMs. To reduce bias, yes/no questions are presented bidirectionally, requiring correct answers in both directions to be considered accurate.}
\vspace{-1.5em}
  \label{fig:framework}
\end{figure*}

The rapid advancement of large vision-language models (LVLMs) has enabled sophisticated tasks involving both visual and textual understanding, such as image and video comprehension. Models like GPT-4o, o3~\cite{openai2024gpt4o}, Gemini~\cite{team2023gemini}, and several open-source video LVLMs~\cite{lin2023video, zhang2024llavanextvideo, wang2024qwen2, zhang2023video, yang2025thinking, xu2024pllava} show strong performance on video understanding and instruction-following tasks. They also perform well on established video benchmarks~\cite{cai2024temporalbench, mangalam2023egoschema, yu2019activitynet, xiao2021next, fu2024video, yu2025unhackable, patraucean2023perception, kesen2023vilma}. However, current benchmarks have notable limitations: many include image-level questions answerable from a single frame~\cite{cai2024temporalbench}, and they often lack diversity in video content and types, limiting a comprehensive assessment of whether LVLMs can truly think based on video. For example, in Figure \ref{fig:teaser}, a video where sequential actions occur (e.g., a person skating around an ice rink, moving out of frame and then back into frame), the model can often answer questions like "What is the person doing?" by identifying the general activity as "skating in circles," but this fails to test true video understanding. For instance, the model cannot determine whether the person moves closer to or farther from the camera first, which requires temporal reasoning across multiple frames.

To address these limitations, this paper proposes \ours\ (Figure \ref{fig:framework}), a new video benchmark designed to evaluate the video understanding capabilities of LVLMs. Compared to other video benchmarks~\cite{li2024mvbench, ning2023video, li2023vlm, liu2024tempcompass}, it leverages the temporal characteristics of 3,269 video data and manually annotates 4,342 high-quality visual-centric question pairs across 11 different scenarios. Specifically, \ours\ challenges models to truly think with videos rather than perform simple understanding, featuring the following characteristics:

\noindent \textbf{Emphasis on deep reasoning content.} Unlike static images, videos contain dynamic information across temporal and spatial dimensions. We manually selected videos with multiple dynamic entities, diverse movements, and complex positional changes that require comprehensive cross-frame analysis. We then constructed questions demanding deep, multi-faceted reasoning rather than superficial pattern recognition. For example, as shown in Figure \ref{fig:framework}, in the "Forensic Authenticity Analysis" category, the question "What suggests algorithmic influence in digital video of natural elements?" requires models to detect subtle inconsistencies in frame transitions and identify artificial backgrounds through detailed analysis. This challenges models to truly think with videos by integrating spatial relationships, temporal sequences, and contextual understanding to uncover complex visual anomalies.

\noindent \textbf{Finer-grained categorization.} Our benchmark spans 11 categories focused on video-specific visual tasks, including trajectory analysis, temporal reasoning, quantitative estimation, event recognition, sequential ordering, and scene context awareness. To enhance coverage, we also include velocity estimation, cinematic dynamics, forensic authenticity analysis, robotics action recognition, and interaction analysis, ensuring \ours\ remains comprehensive for evaluating text-to-video generation and embodied environments.

In addition to these major characteristics, to further facilitate automated evaluation and reduce biases during the assessment process, we structured the questions in a multiple-choice format. For yes/no-type questions, we created paired questions by reversing the order of actions or subjects. For example, "Does the person in the video open the cabinet before turning on the light? Answer: yes" is paired with "Does the person in the video turn on the light before opening the cabinet? Answer: no". The model is considered correct only when it answers both questions in the pair accurately.

Using \ours, we conducted a comprehensive evaluation of several state-of-the-art LVLMs, including GPT-4o, o3~\cite{openai2024gpt4o}, Gemini-1.5~\cite{team2024gemini}, along with open sourece LVLMs like VideoLLaVA~\cite{lin2023video}, LLaVA-NeXT-Video~\cite{zhang2024llavanextvideo}, Video-LLaMA~\cite{zhang2023video}, Video-LLaMA2~\cite{cheng2024videollama},Chat-UniVi-V1.5~\cite{jin2024chat} and Qwen2-vl~\cite{wang2024qwen2}. As a comparison, we also conducted experiments on LVLMs~\cite{liu2024improved, ye2024mplug, bai2023qwen} fine-tuned on image-based instruction data to demonstrate that the tasks in \ours\ cannot be effectively answered using single-frame images. We found that the most advanced LVLMs, Gemini-1.5 pro~\cite{team2024gemini}, achieved an accuracy of only 56.98\% on \ours, significantly lower than the average score 94.82\% of human volunteers on a randomly sampled subset. Among the open-source models fine-tuned on video datasets, the best-performing model, Qwen2-VL, also achieved only 52.47\% accuracy. Furthermore, the best-performing image-based LVLMs, LLAVA-1.5, achieved an accuracy of 37.48\%, indicating that relying solely on single-frame or multi-frame image data in a single modality is insufficient for effectively utilizing and capturing the temporal information in video data. Additionally, we observed that LVLMs perform worse on videos with longer average lengths, highlighting that current LVLMs struggle to handle long video sequences effectively. The issues exposed by these findings provide guidance for the subsequent optimization of the model.

\section{Related Work}
\label{sec:related_work}
\vspace{-0.5em}
\textbf{Large Vision Language Models.} Large language models (LLMs)~\cite{openai2023b, touvron2023llama, alpaca, vicuna2023} have demonstrated impressive text comprehension capabilities. With the integration of visual components~\cite{ye2024mplug, zhu2023minigpt, liu2023llava, qutexttt, zhou2024calibrated, wang2024enhancing}, these text models have gained the ability to understand multimodal data, allowing them to interpret real-world images. Notable examples include commercial models like GPT-4V~\cite{openai2023gpt4v} and Gemini-1.5~\cite{team2024gemini}, as well as open-source models such as LLaVA-1.5~\cite{liu2024improved}, Qwen-VL~\cite{bai2023qwen}, and mPlug-Owl~\cite{ye2023mplug}. The development of LVLMs has expanded their capabilities from static image understanding to complex video comprehension. Early work, such as VideoChat~\cite{li2023videochat} and Video-ChatGPT~\cite{maaz2023video}, integrated visual encoders with language models, laying the foundation for video-based multimodal dialogue. Subsequent research introduced improvements~\cite{zhang2023video, jin2023chatunivi, ren2024timechat, song2024moviechat, liu2023video}, including the addition of audio modalities, joint training on images and videos, and optimization of feature alignment.

\begin{figure}[t]
  \centering
  \includegraphics[width=0.44\textwidth]{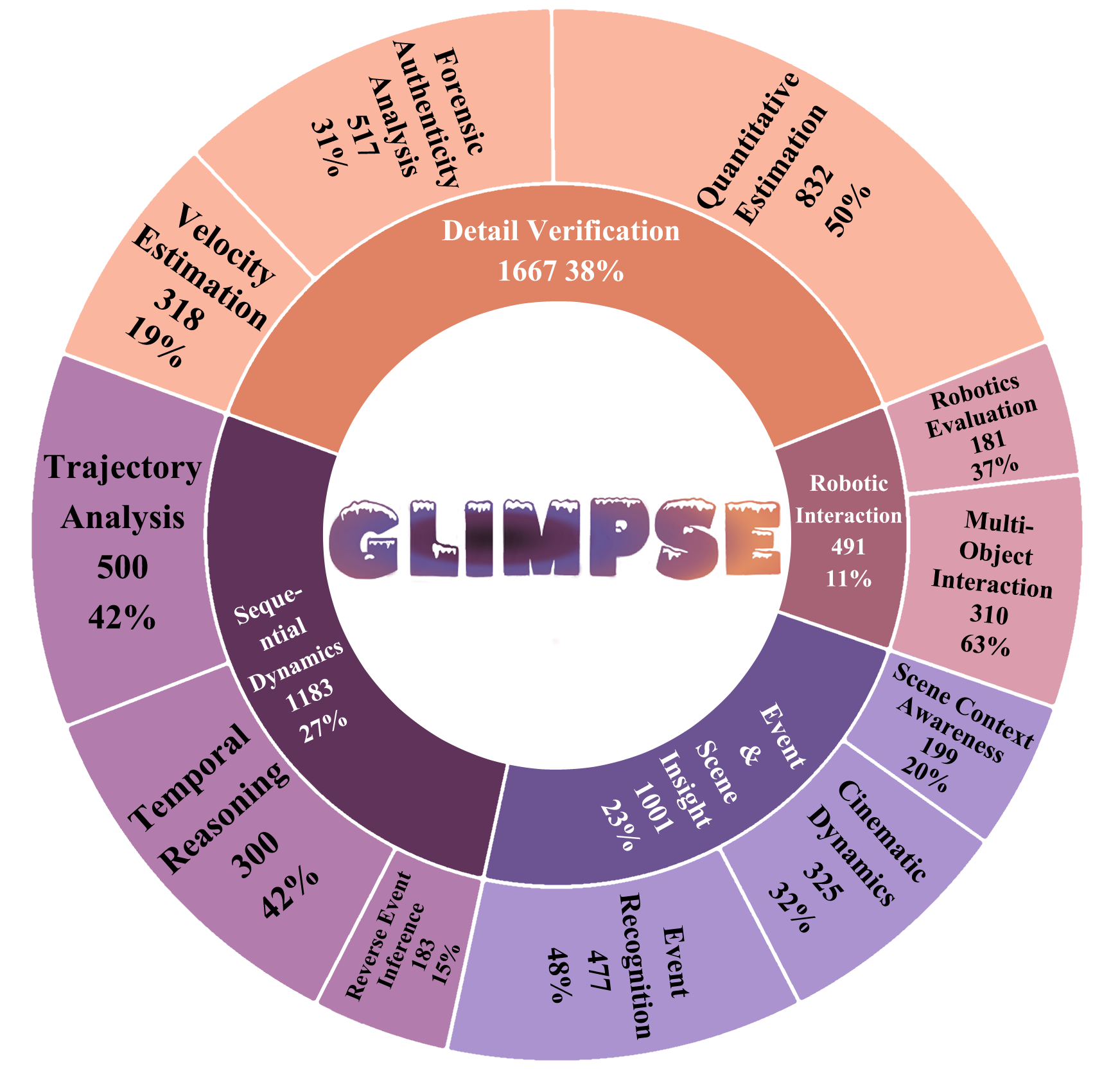}
  \vspace{-10pt}
\caption{Distribution of categories in the \ours\ benchmark. Our benchmark covers 4 key domains and 11 detailed visual-centric question types.}
\vspace{-1.5em}
  \label{fig:category}
\end{figure}

\noindent\textbf{Video Understanding Benchmarks.}  Benchmarking efforts have increasingly extended from static images to the video domain, where temporal dynamics and content complexity demand new evaluation approaches. Early work like SEED-Bench~\cite{li2023seed} supports multimodal evaluation for both image and video QA, including temporal modeling dimensions. AutoEval-Video~\cite{chen2023autoeval} and Video-Bench~\cite{ning2023video} are specifically designed for video scenarios and generate video QA data using LLMs. MVBench~\cite{li2024mvbench} introduces an innovative approach by repurposing existing datasets for LVLMs evaluation. ET-Bench~\cite{liu2024bench} targets complex, multi-event, time-sensitive tasks to assess temporal and contextual understanding. Despite this progress, no existing benchmark specifically targets high-quality, vision-centric video QA. To address this gap, this paper introduces \ours, a benchmark focused on evaluating LVLMs' visual perception and reasoning in video contexts.

\section{The GLIMPSE Benchmark}
\label{sec:GLIMPSE}
\vspace{-0.5em}

\subsection{Overview}
In this section, we introduce \ours, a benchmark focused on visual-centric content. By "visual-centric," we refer to content where understanding requires comprehensive analysis of dynamic visual elements across multiple frames, such as tracking object movements, analyzing complex interactions, and interpreting evolving spatial arrangements, rather than relying on superficial glimpses or single-frame observations. This benchmark challenges models to truly think with videos through sustained analytical engagement. 
As shown in Figures~\ref{fig:framework} and ~\ref{fig:category}, \ours\ comprises a total of 3,269 carefully selected video instances and 4,342 unique questions, all manually constructed. Answering each question requires a comprehensive understanding of the video. Based on the temporal characteristics of video data, we categorize the questions into 11 subcategories.

\begin{figure}[t]
  \centering
  \includegraphics[width=0.45\textwidth]{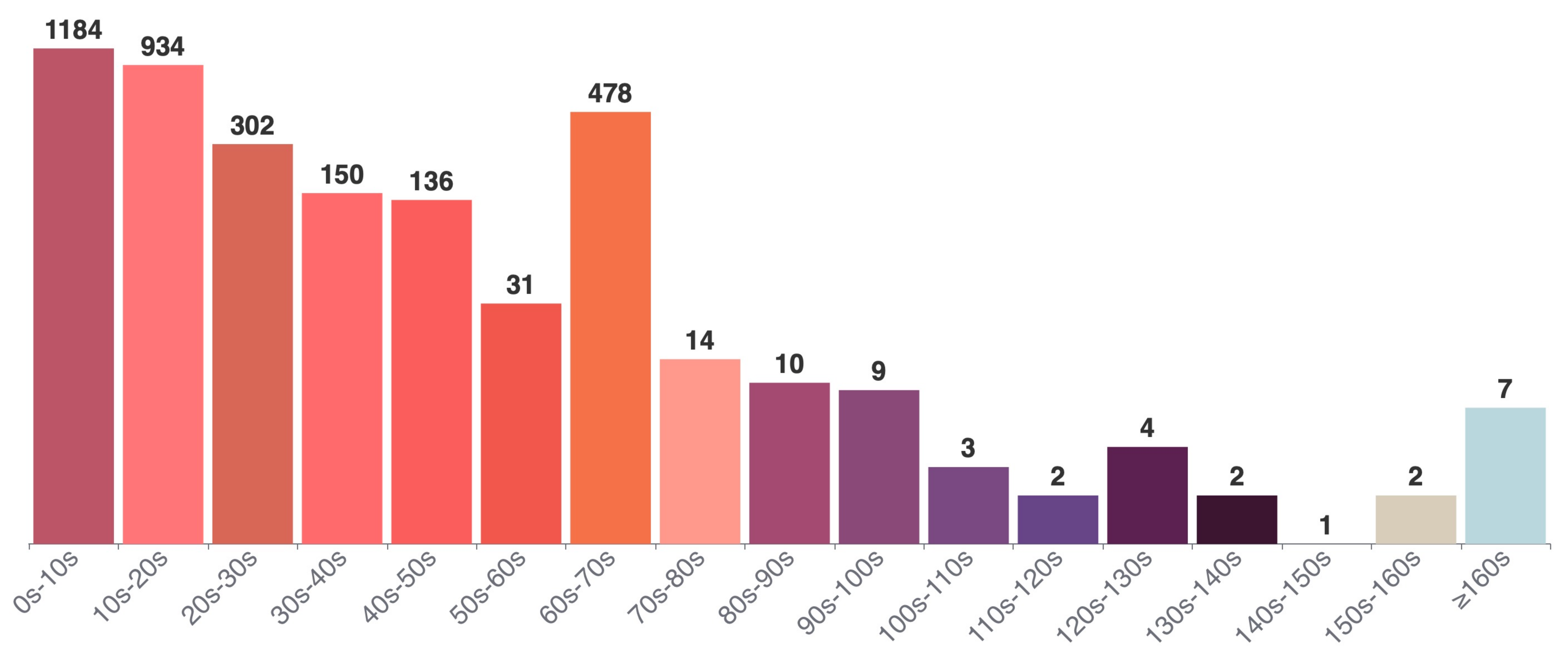}
  \vspace{-10pt}
\caption{Video length distribution, with the horizontal axis showing duration and the vertical axis indicating the number of videos per range.}
\vspace{-15pt}
  \label{fig:len}
\end{figure}

\subsection{Dataset Curation}
The data curation process of the \ours\ dataset consists of three key steps: video collection, question-answer annotation, and quality review. Each step is designed to ensure comprehensive coverage and high-quality representation across various categories in our benchmark, while also making certain that each question is visual-centric and closely aligned with the benchmark's theme. We detail the process as follows:

\noindent\textbf{Video Collection.} To annotate visual-centric QA pairs, we first select a comprehensive set of dynamic videos covering a wide range of types. We began by identifying the visual content of interest in each video, categorizing it into 11 key areas: \textbf{(1) Trajectory Analysis:} This task focuses on analyzing the trajectory or path of objects within the video, involving an understanding of movement direction, displacement over time, and overall motion patterns. The model needs to comprehend the video, identify the objects to be tracked, and make integrated judgments based on the video's context. This helps evaluate the model's fine-grained recognition and temporal reasoning abilities. \textbf{(2) Temporal Reasoning:} Involves understanding the timing and sequence of events. Although previous benchmarks, such as~\cite{fu2024video, li2024mvbench, liu2024bench, ning2023video}, have explored similar tasks, they mostly focus on locating a single action, object, or event (e.g., asking when an object appears). In contrast, \ours\ aims to provide a more visually-centric evaluation by asking about the temporal order of events, such as whether a specific event occurred before another. This approach effectively assesses the model's temporal reasoning ability. \textbf{(3) Quantitative Estimation:} This task involves estimating quantities such as distances or counts within the video, adapted from quantitative tasks in image modalities. In \ours, we focus on dynamic events—for example, counting how many times a person performs a specific action or how many times an animal appears and disappears. An example question could be, "How many times did the dog and the person clap hands in the video?" This type of question emphasizes the model’s ability to understand and track dynamic information in video content. \textbf{(4) Event Recognition:} This task aims to determine whether events occur in the video and their sequence relative to other events, assessing the model's understanding of video content and temporal relationships, especially in scenarios where multiple events happen sequentially. The model must accurately identify events and the logical order between them. \textbf{(5) Reverse Event Inference:} Determines the correct sequence of events or actions, reconstructing the event flow from partial information. \textbf{(6) Scene Context Awareness:} Understands changes in the background of the scene in the video (e.g., location or setting). This task evaluates the model's spatial understanding and context recognition abilities. For example, "How do the traffic lights change throughout the video?" \textbf{(7) Velocity Estimation:} Calculates the relative speed of moving objects, analyzing their displacement over time. For example, "How does the speed of the black and white horses change?" \textbf{(8) Cinematic Dynamics:} This task focuses on identifying camera motion in video footage, presenting a unique challenge. The model needs to fully understand both the foreground and background of the video and analyze their movement in relation to each other to determine the style of camera motion. For example, "How does the camera move throughout the video?" \textbf{(9) Forensic Authenticity Analysis:} By collecting and generating some fake video data using text-to-video models~\cite{podell2023sdxl}, we can then use this data for evaluation to test whether current models can identify fake videos. This approach effectively verifies the model’s ability in assessing video authenticity. \textbf{(10) Robotics Evaluation:} Identifies actions performed by robots, such as grasping, moving, and assembling, to assess the stability, success, and effectiveness of various robotic tasks. \textbf{(11) Multi-Object Interaction:} Focuses on analyzing the interactions between multiple objects or entities (such as robots, people, animals, or specific items) within a given scenario, including actions like physical contact, collaboration, or conflict. For instance, "What did the person do with the box when they approached it in the final scene of the video?"

After defining the categories, we carefully selected and collected high-quality video resources that meet the specific requirements of each category to ensure comprehensive coverage across the benchmark. For event recognition, we extracted and reconstructed data from ShareGPTVideo~\cite{chen2024sharegpt4video}. Robotics action recognition included videos sourced from the PushT dataset~\cite{chi2024diffusionpolicy} and Cable Routing~\cite{luo2023multistage}. For temporal reasoning, we chose videos from Ego4D~\cite{grauman2022ego4d} and Pexels\footnote{\url{https://www.pexels.com/}}. Quantitative estimation and velocity estimation tasks utilized selected videos from Panda-70m~\cite{chen2024panda}. The remaining videos were curated from Pexels and Pixabay\footnote{\url{https://pixabay.com/}}. Additionally, to avoid excessive complexity in the testing tasks and maintain reference value, we controlled the length of the selected videos to be between 20 seconds and 2 minutes. The final dataset includes 3,269 videos with a balanced distribution of video lengths, as shown in Figure \ref{fig:len}.

\noindent\textbf{Question-answer Annotation.} After collecting the raw video data, we manually annotated high-quality question-answer pairs to evaluate the ability of LVLMs in interpreting video content. To facilitate the annotation process, we enlisted researchers proficient in English to create open-ended questions and answers. Specifically, the annotators first watched the entire video and, by re-watching as needed, generated 1–3 questions per video. For easier testing, we then used GPT-4o API to convert these open-ended question-answer annotations into a multiple-choice format suitable for automated evaluation. For yes/no-type questions, such as those in the "temporal reasoning" category, we aimed to reduce bias during evaluation (e.g., random guessing with a 50\% accuracy rate). To address this, we used the GPT API to construct bidirectional question pairs, as exemplified in the "temporal reasoning" samples shown in Figure \ref{fig:framework}. For example, we initially manually annotated the question-answer pair as: Q1: "A man first took out a paper box, then put the phone back in his pocket." A) Yes B) No. Then we used GPT-4o API to modify it into a reverse question-answer pair: Q2: "Did a man put the phone back in his pocket before taking out a paper box?" A) Yes B) No. The specific prompts used can be found in Appendix \ref{sec:ap_prompt}. Only when LVLMs answered both questions correctly was the response considered accurate. 

\noindent\textbf{Quality Review.} To ensure the quality of the dataset and confirm that the manually constructed questions and selected videos are indeed visual-centric, we implemented a rigorous manual review process. First, different annotators were assigned to check each question-answer pair to ensure that each question is well-defined and answerable based on the video content. For example, questions like "What is the approximate speed of the vehicle in the video?" lack a clear answer and are removed during screening. We also ensured that each question required understanding of the entire video rather than a single frame. For example, questions like "What is the weather in the video?" could be answered with just one frame, so similar questions were filtered out during the review process. Through annotation, manual selection and quality review, we ultimately collected a total of 4342 high-quality question-answer pairs. Detailed examples can be found in Appendix \ref{sec:ap_case}.

\section{Experiment}
\label{sec:Experiment}
\vspace{-0.5em}

\begin{table*}[t!]
\centering
\footnotesize
\vspace{-10pt}
\resizebox{\textwidth}{!}{
\begin{tabular}{l|cccccccccccc}
\toprule
\multicolumn{1}{c|}{\textbf{Models}} & \textbf{TA} & \textbf{TR} & \textbf{QE} & \textbf{ER} & \textbf{REI} & \textbf{SCA} & \textbf{VE} & \textbf{CD} & \textbf{FAA} & \textbf{RE} & \textbf{MOI} & \textbf{Avg} \\
\midrule
Random & 23.60 & 25.40 & 23.92 & 24.32  & 16.39  & 25.63  & 25.18  & 24.31 & 27.08 & 23.53  & 22.58   & 24.14 \\
\midrule
\multicolumn{13}{l}{\textit{Open-source Image LVLMs: All models use a randomly sampled single frame as input.}}\\
\midrule
mPLUG-OWL2 (7B) & 39.25 & 22.40 & 36.90 &31.80  &34.00  &37.30  &28.60  & 38.40 &52.70 &33.50  &28.34    & 34.12 \\
Qwen-VL-Chat (7B) & 38.24 & 11.60  & 35.71 &30.72   &32.79  &36.18  &27.48  &33.16  &51.62  &32.04  &27.42  & 30.73  \\
LLAVA-1.5 (7B) &42.03 &28.47 &30.02 &42.96 &45.53 &28.52 &29.48 &41.47 &67.04 &30.01 &27.03 &37.48\\

\midrule
\multicolumn{13}{l}{\textit{Open-source Video LVLMs: All models use their default numbers of frames as inputs.}}\\
\midrule
Video-LLaMA (7B) &44.71 &28.88& 23.30& 48.94& 38.00& 60.46& 15.32& 53.30& 65.09& 42.13& 51.16  & 39.71  \\
Video-LLaMA2 (7B) &47.06  &30.40  &24.53  &51.52  & 40.00 &63.64  &16.13  &56.10  &68.52  & 65.19 & 53.85 & 42.60 \\
Chat-UniVi-V1.5 (7B) &39.86  &23.68  &22.36 &54.10& 42.00& 66.82& 16.94& \textbf{58.91} &71.95& 68.45& 56.54  & 41.47  \\
LLAVA-NeXT-Video (7B) & 46.79 & 42.33 & 19.55 & 57.68 & 44.45 & 69.04 & 34.48 & 53.03 & 72.87 & 63.71 & 57.17 & 46.80  \\
VideoLLaVA (7B) & 40.42 & 22.56 & 21.86 & 52.60 & 42.99 & 68.00 & 17.60 & 56.48 & 71.39 & 65.23 & 54.95 & 40.74  \\
Qwen2-VL (7B) & 46.15 & \underline{44.44}  &28.37  & 59.32  & 43.42 & 67.84  &33.18  & 55.32 & 73.44 & 66.85 & 58.82 & 52.47 \\

\midrule
\multicolumn{13}{l}{\textit{Closed-source LVLMs: All models use fixed interval sampling.}} \\
\midrule
GPT-4o &48.40  &28.80  &49.64  &59.12  &56.83  & 62.81&40.88  & 52.92 & 65.18 &70.11  & 57.10 &53.80  \\
GPT-o3 &\underline{55.42}  &\textbf{53.07}  &\textbf{65.75}  &61.51  & \textbf{67.39}  & \textbf{82.00}& \textbf{62.90}  & \underline{56.23} & \textbf{85.69} &69.55  & \textbf{71.20} &\textbf{66.43}  \\
Gemini 1.5 Flash & 54.60 & 33.60   & \underline{64.90}  & \underline{62.68}  & 54.10  & 69.00  & 44.14  & 51.38  & 73.11  & \underline{73.48}  & 61.62  & 55.65  \\
Gemini 1.5 Pro & \textbf{61.02}  & 42.84  & 51.32  & \textbf{64.10} & \underline{56.86} & \underline{72.12} & \underline{45.45} & 53.97 & \underline{75.24} & \textbf{77.95}  & \underline{62.64} & \underline{56.98} \\
\midrule
\multicolumn{13}{l}{\textit{Human Performance}}\\
\midrule
Human Expert & 92.00 & 96.00 & 88.00 & 100.0 & 92.00 & 100.0 & 94.00 & 96.0 & 91.00 & 96.00 & 98.00 & 94.82 \\
\bottomrule
\end{tabular}
}
\caption{The overall performance of representative LVLMs on the \ours\ benchmark across various categories. Here, we report accuracy for each specific category in terms of visual-centric question answering. Specifically, \textbf{TA} denotes Trajectory Analysis, \textbf{TR} denotes Temporal Reasoning, \textbf{QE} denotes Quantitative Estimation, \textbf{ER} denotes Event Recognition, \textbf{REI} denotes Reverse Event Inference, \textbf{SCA} denotes Scene Context Awareness, \textbf{VE} denotes Velocity Estimation, \textbf{CD} denotes Cinematic Dynamics, \textbf{FAA} denotes Forensic Authenticity Analysis, \textbf{RE} denotes Robotics Estimation, and \textbf{MOI} denotes Multi-Object Interaction. In each column, the best performance, excluding human experts, is bolded, and the second-best is underlined.}
\label{sec:Experiment_tabel}
\vspace{-15pt}
\end{table*}
\ours\ systematically evaluates the understanding and perception capabilities of existing LVLMs on video data. In this section, we aim to address the following key questions: (1) Can current LVLMs truly think with video content and answer questions accurately? (2) How large is the gap between current LVLMs and human performance in understanding video content? (3) Do the evaluated LVLMs show a preference for a certain field?

\subsection{Experiment Setup}
\textbf{Baseline Models.} We first tested (1) three commercial LVLMs: GPT-4o, o3~\cite{openai2024gpt4o}, Gemini 1.5 Flash, and Gemini 1.5 Pro~\cite{team2024gemini} on the \ours\ benchmark; (2) evaluated representative video-based LVLMs, including Video-LLaMA~\cite{zhang2023video}, Video-LLaMA2~\cite{cheng2024videollama}, LLaVA-NeXT-Video~\cite{zhang2024llavanextvideo}, VideoLLaVA~\cite{lin2023video}, Qwen2-VL~\cite{wang2024qwen2}, and Chat-UniVi-V1.5~\cite{jin2024chat}; and (3) to examine the differences between image LVLMs and video-based LVLMs, we also tested representative models such as LLAVA 1.5~\cite{liu2024improved}, Qwen-VL~\cite{bai2023qwen}, and MPlug-Owl 2~\cite{ye2024mplug}. For video-based models, we followed the frame sampling methods provided by the respective authors during evaluation. For image-based LVLMs, we randomly sampled a single frame as input. For GPT-4o, o3, Gemini-1.5 Flash and Gemini-1.5 Pro, we used fixed interval sampling with parameters set to sample\_frequency = 50 and max\_frame\_num = 16.

\noindent \textbf{Evaluation Metrics.} We use accuracy to evaluate the performance of LVLMs on \ours, where for Temporal Reasoning, questions are constructed in a bidirectional format—meaning each video has two related questions. Thus, when calculating the accuracy for this category, we only count a response as correct if the model answers both questions correctly for a given video.

\noindent \textbf{Human Evaluation.} For comparison, we randomly selected 50 questions from each of the 11 classes in \ours, forming a 550-sample subset, and invited five student volunteers to answer them. Volunteers were instructed to watch each video once without replay and answer immediately. They also recorded the time spent per question (excluding video watching). The total time was 1327 seconds, averaging 2.4 seconds per question.

\subsection{Quantitative Results}
This section presents the evaluation results on \ours\, with detailed LVLM performance in Table \ref{sec:Experiment_tabel}. To provide a comparison baseline, we include a "Random" row, illustrating task difficulty and whether models perform above chance. "Random" results are computed by generating equal numbers of random A/B/C/D or Yes/No answers based on total question count. Based on Table~\ref{sec:Experiment_tabel} and human performance, our key findings are:

\noindent\textbf{Gap between video-based LVLMs and image-based LVLMs.} We found that image-finetuned models like LLAVA-1.5, mPLUG-OWL2, and Qwen-VL-Chat, while performing well on existing image benchmarks, showed weaker overall performance on the \ours\ benchmark compared to LVLMs fine-tuned on video data. This gap is especially noticeable in tasks requiring video context understanding, such as Trajectory Analysis, Temporal Reasoning, Quantitative Estimation, and Sequential Ordering. For example, in the typically lowest-scoring task of Temporal Reasoning, the best-performing model, GPT-o3, scored 53.07, while the top image LVLMs, LLAVA-1.5, only achieved 28.47. Additionally, video LVLMs outperformed image LVLMs in this task by an average of 56.09\% in accuracy. These findings indicate that specialized training on high-quality video data enhances model performance in tasks with high temporal requirements. The results also indicate that single-frame images lack sufficient information to answer questions in \ours. Improving visual components and developing methods to extract comprehensive and temporal video features may be crucial for further advancing video-based LVLMs performance.

\noindent\textbf{Challenges of Current LVLMs in Temporal Understanding and Fine-Grained Visual Tasks.} Based on the results in Table \ref{sec:Experiment}, it can be seen that although GPT-o3 performed the best on the \ours\ benchmark, its overall average score was only 66.43, significantly below human-level performance. Specifically, in challenging tasks within \ours\ such as Temporal Reasoning, Trajectory Analysis, and Quantitative Estimation, the overall performance of the tested models was generally modest. For Temporal Reasoning, the top model, GPT-o3, achieved only 53.07 accuracy, while other models were close to random performance levels. In Quantitative Estimation, apart from GPT-o3, which achieved 65.75 accuracy, other models also showed average performance. This indicates that current LVLMs still face significant challenges in understanding temporal information in videos and in fine-grained visual-centric tasks.

\noindent\textbf{Image Understanding Capabilities Are Equally Important for Enhancing Video Understanding.}
In open-source video LVLMs, most models outperform open-source image LVLMs overall, especially in tasks requiring temporal information. However, in tasks such as Quantitative Estimation and Velocity Estimation, video LVLMs generally perform worse than image LVLMs and closed-source commercial LVLMs. This discrepancy may be due to video LVLMs focusing more on temporal and dynamic features and lacking sufficient optimization and fine-grained data to handle static quantitative estimation and speed calculation tasks, impacting their performance in these specific areas.

\subsection{Time-Related Performance Analysis}

Due to the varying amounts of information covered by videos of different lengths, we conducted an in-depth exploration of the impact of video length on model performance. We selected two categories with relatively even distributions of video duration: Temporal Reasoning and Scene Context Awareness, and tested three models from different categories in Table \ref{sec:Experiment}: GPT-4o, Qwen2-VL, and LLAVA-1.5. We divided the original questions into five subsets based on 10-second intervals and tested these models on each subset. The test settings were the same as in Table \ref{sec:Experiment}. We calculated the accuracy of these models for questions corresponding to videos within each time interval, and the results are shown in Figure \ref{fig:len_cp}. From the results, we can make the following observations:
\begin{figure}[t]
  \centering
  \includegraphics[width=0.45\textwidth]{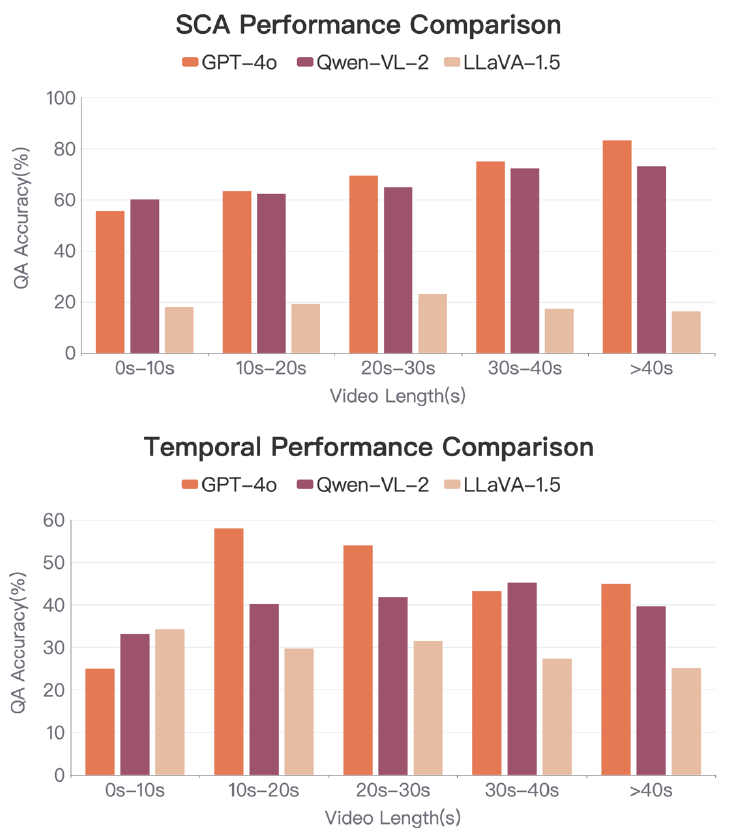}
  \vspace{-10pt}
\caption{Model performance comparison in Scene Context Awareness (SCA) and Temporal Reasoning across different video durations. The y-axis shows accuracy, and the x-axis shows time intervals.}
\vspace{-17pt}
  \label{fig:len_cp}
\end{figure}

\begin{table}[ht]
    \centering
    \resizebox{\linewidth}{!}{
    \begin{tabular}{lccc}
        \hline
        \textbf{Video Length} & \textbf{GPT-4o} & \textbf{Qwen2-VL} & \textbf{LLAVA-1.5} \\
        \hline
        0--10s  & 42.33 & 34.67 & 30.00 \\
        10--20s & 48.67 & 42.33 & 38.33 \\
        20--30s & 55.00 & 48.00 & 45.00 \\
        30--40s & 58.33 & 52.00 & 37.67 \\
        $>$40s  & 64.67 & 56.00 & 32.00 \\
        \hline
    \end{tabular}
    }
    \caption{The overall performance of models on different video lengths across 12 categories.}
    \label{tab:video_length_comparison}
\end{table}

\noindent \textbf{More information helps improve performance.} For Scene Context Awareness tasks, the accuracy of the model's responses significantly improves as video length increases. This is because longer videos provide more information, allowing the model to capture richer background details and environmental context. This additional contextual information helps the model build a more comprehensive understanding of the scene, including objects, spatial relationships, and the overall atmosphere. With more time to "observe" and "learn" the nuances within the video, the model can more accurately grasp scene characteristics, making it easier to generate responses that reflect the actual context. This indicates that video length directly impacts model performance in scene understanding tasks, suggesting that long video understanding will be an important area of research in the future.

\noindent \textbf{Challenges Remain in Processing Long Videos.} For Temporal Reasoning tasks, model accuracy initially increases with video length but then declines. Short videos have limited time spans, and fewer frames are input, making it difficult for the model to capture enough information for accurate reasoning. As the video length increases to between 20 and 30 seconds, the additional information helps the model better understand temporal cues, thus improving reasoning accuracy. However, Temporal Reasoning is inherently challenging; as video length continues to exceed 30 seconds, the complexity and time span of events increase, dispersing the information. This makes it difficult for the model to effectively process long videos, leading to a drop in performance.

\begin{figure}[t]
  \centering
  \includegraphics[width=0.46\textwidth]{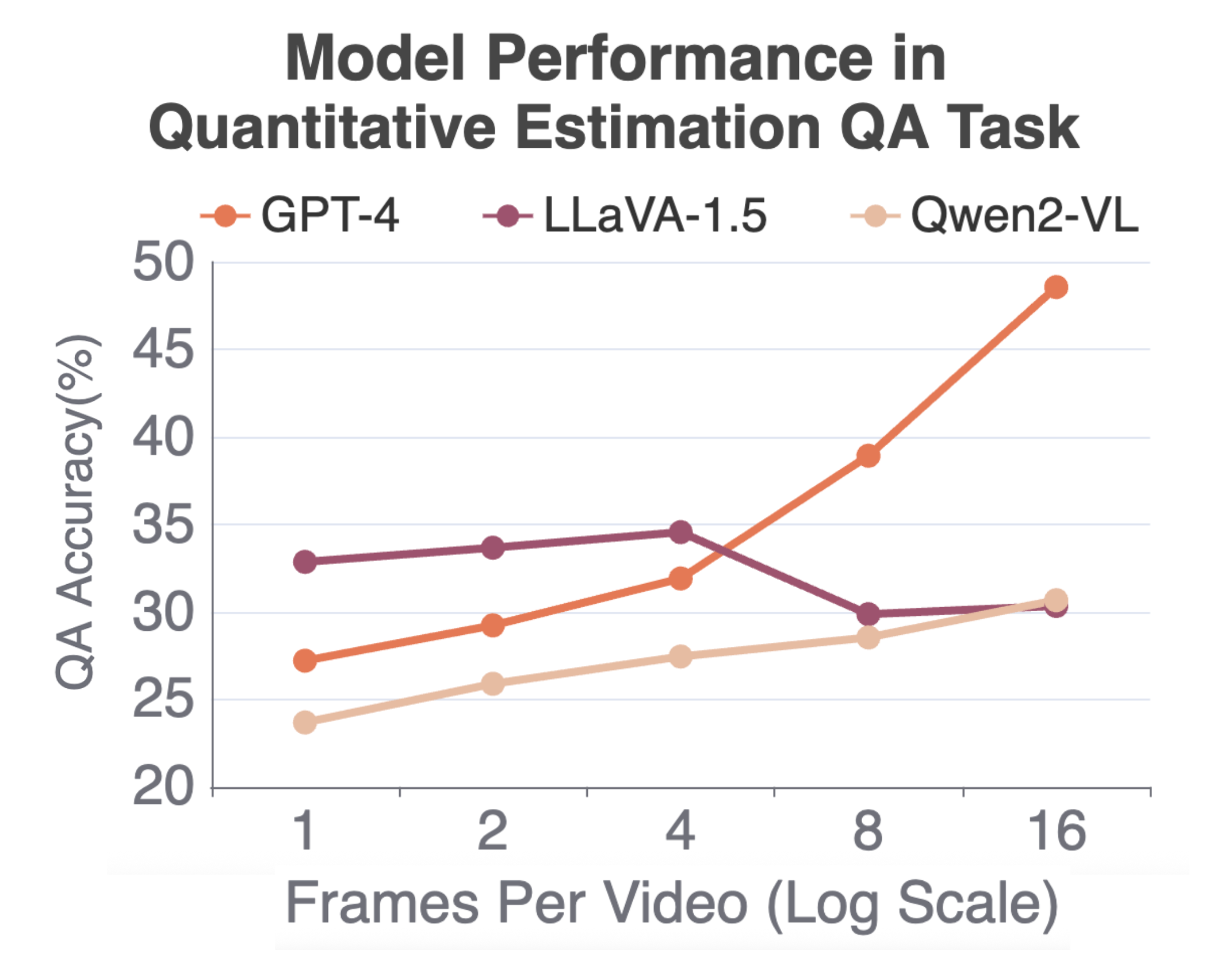}
  \vspace{-10pt}
\caption{Model performance on quantitative estimation with varying frame counts.}
\vspace{-10pt}
  \label{fig:frame}
\end{figure}

\begin{figure}[t]
  \centering
  \includegraphics[width=0.46\textwidth]{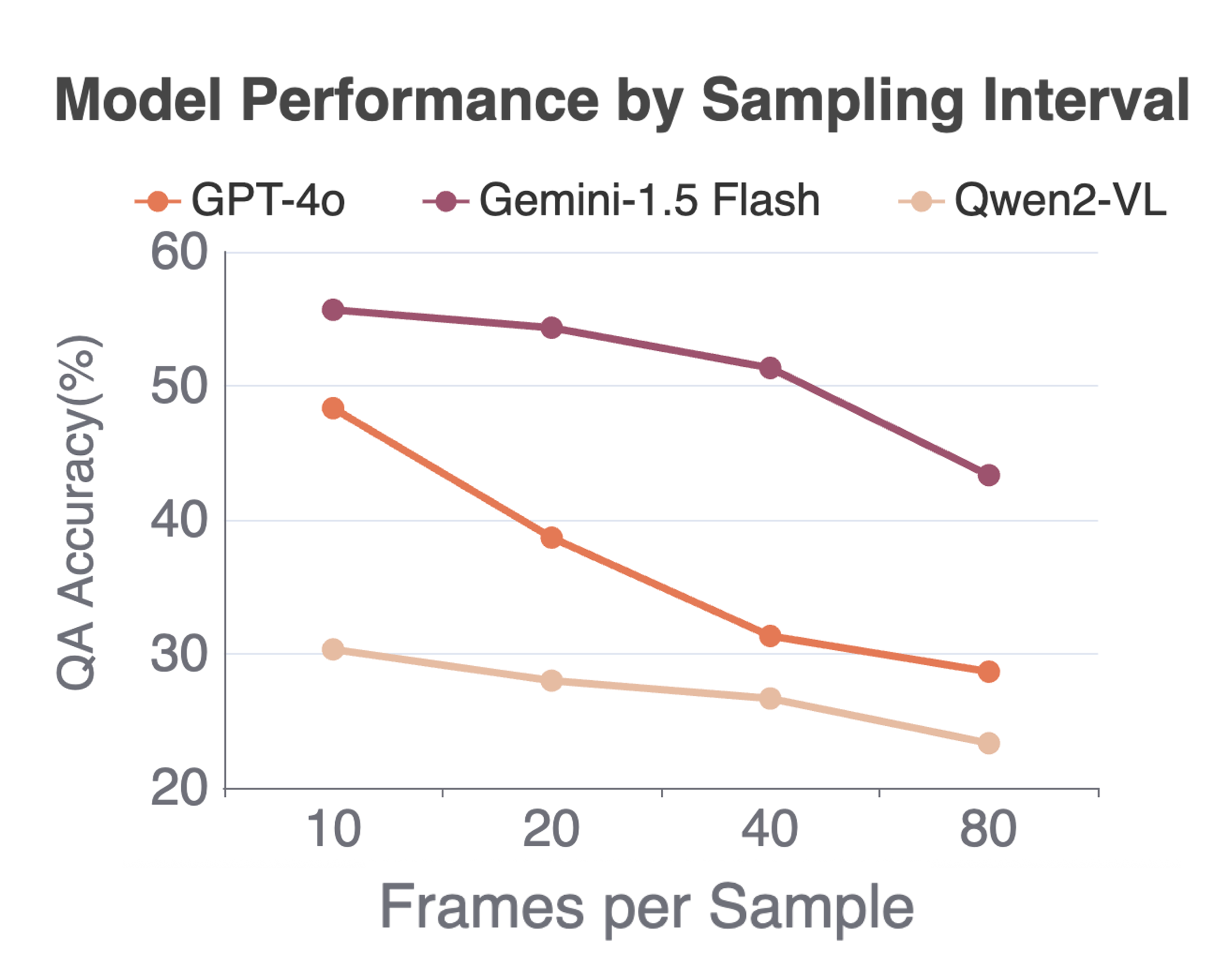}
  \vspace{-10pt}
\caption{Model performance of Every N Frames sampling.}
\vspace{-17pt}
  \label{fig:frame_sampling}
\end{figure}

\subsection{Effect of the Number of Input Frames}
In this section, we analyze how frame count affects model performance, focusing on the Quantitative Estimation task. This task is especially sensitive to the amount of visual information, making it well-suited for examining the impact of varying frame counts. Unlike tasks that rely on temporal or sequential understanding, Quantitative Estimation requires detailed snapshots of individual frames, as these provide data points for counting or assessing quantities. By analyzing model performance on this task, we can more clearly observe how frame count affects the model’s ability to interpret fine-grained visual information without the added complexity of temporal dependencies. We tested three models—GPT-4o, Qwen2-VL, and LLAVA-1.5—using 1, 2, 4, 8, and 16 frames as input, and plotted their performance as frame count increases. The experimental results are shown in Figure \ref{fig:frame}. We observe that initially, increasing the frame count improves the accuracy of all models; however, once the frame count surpasses a certain threshold, performance gains diminish and even start to decline. This suggests that, in the Quantitative Estimation task, a moderate number of frames enhances model accuracy, as more frames capture additional key details related to quantity. However, further increasing the frame count leads to a drop in performance, possibly because the model struggles to process excessive frame information effectively, resulting in redundancy and difficulty focusing on relevant features.

\begin{figure*}[t]
  \centering
  \includegraphics[width=0.87\textwidth]{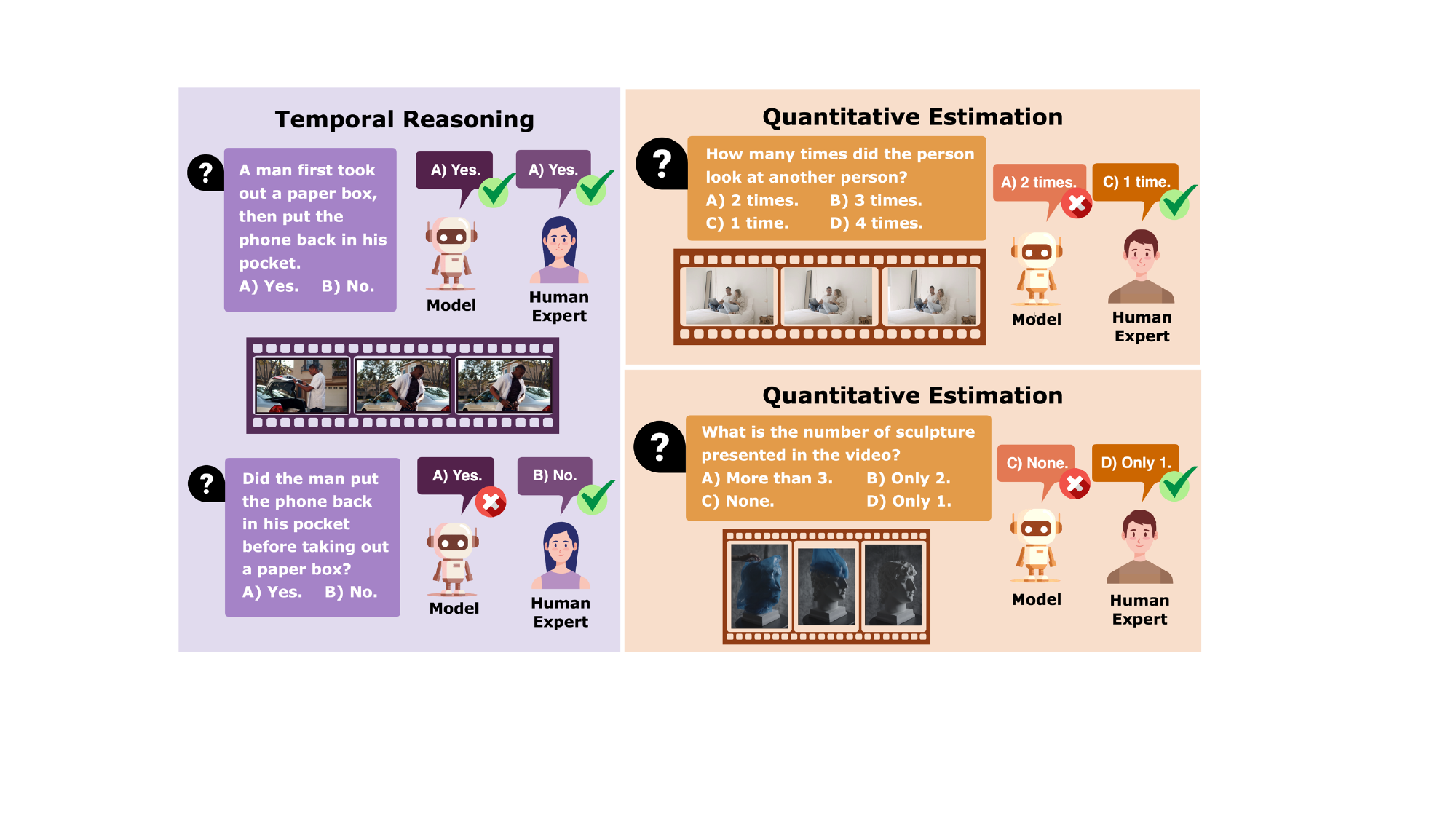}
  \vspace{-8pt}
\caption{The failure cases of GPT-4o in \ours. It can be observed that GPT-4o has difficulty accurately understanding quantity-related questions in the video. Additionally, when answering temporal reasoning questions, it makes mistakes when the question is reversed.}
\vspace{-1em}
  \label{case}
\end{figure*}
\section{Error Analysis}
\label{sec:Analysis}
This section provides an in-depth error analysis, focusing on two tasks with lower average accuracy: Temporal Reasoning and Quantitative Estimation, which are easy for humans but error-prone for models. Using GPT-4o and Gemini 1.5 Flash as examples, we present case studies in Figure~\ref{case}.

\subsection{Temporal Reasoning}
In our experiments, we found that the Temporal Reasoning task had the highest error rate. For this task, to reduce potential biases, we structured each temporal reasoning question to include contrasting statements, such as "Question: Does Event A occur before Event B? Answer: yes" and "Question: Does Event A occur after Event B? Answer: no". Only when both answers in a question pair were correct was the response considered accurate. Here, we refer to questions with a "yes" answer as positive questions. Under this setup, we found that when evaluating GPT-4o, if we assessed its accuracy on positive questions alone, it achieved 52.30\%, but its accuracy on question pairs was only 28.80\%. The primary source of errors in question pairs was an inconsistent selection pattern, where one answer was correct and the other incorrect. An example of this is shown on the right side of Figure \ref{case}. This suggests that GPT-4o does not truly understand and integrate the video content and question context when answering, as it may struggle to establish the correct temporal sequence during video encoding and frame extraction, thereby impacting its understanding of event order and sequence. It highlights that \ours\ can effectively reduce bias in the evaluation process by constructing bidirectional question-and-answer formats.

\subsection{Quantitative Estimation}
In the Quantitative Estimation task, there are two main issues: quantity errors and bias toward certain options. Using Gemini-1.5 Flash as an example, our investigation reveals that 62\% of errors in Quantitative Estimation task occur when the model selects options like "None" or "No object", leading to incorrect quantity estimation. For instance, as shown in the example on the left of Figure \ref{case}, when asked how many sculptures were present in the video, the model responded with "None", despite a stationary sculpture being visible throughout the video. In addition, quantity errors are also significant. For example, in a question about the number of times a man looked at a woman in the video, GPT-4o counted each instance of the man turning his head toward the woman and then back as two separate actions, resulting in an inflated count. This type of quantity error may also stem from the model's difficulty in accurately understanding fine-grained, visually-centric information.

\section{Conclusion}
\label{sec:Conclusion}
\vspace{-0.5em}

This paper presents \ours, a benchmark dataset for evaluating large vision-language models (LVLMs) on vision-centric video understanding and reasoning. \ours\ targets key skills such as trajectory tracking, temporal reasoning, quantity estimation, scene comprehension, and interaction analysis. It features diverse video content and carefully designed question pairs to comprehensively assess models’ perceptual and reasoning abilities, as well as whether they can truly think based on video. To ensure accurate and scalable evaluation, \ours\ adopts multiple-choice and bidirectional QA formats that mitigate assessment bias. Experimental results reveal that even state-of-the-art multimodal models lag far behind human performance on \ours, underscoring the challenges and opportunities in deep video understanding.

\section*{Limitations}

While \ours\ advances the evaluation of LVLMs in video understanding, it has several limitations. First, the dataset's focus on pre-selected video categories may underrepresent niche or unconventional scenarios, limiting generalizability. Second, restricting videos to 20 seconds–2 minutes overlooks challenges in ultra-long or extremely short contexts. Additionally, the multiple-choice format simplifies real-world open-ended reasoning, narrowing the assessment of nuanced understanding. Human annotation, though rigorous, introduces potential biases in question design and answer validation. Finally, benchmarked models may not fully encompass emerging or specialized LVLM architectures. Addressing these limitations could improve dataset diversity, evaluation flexibility, and model inclusivity in future work.

\bibliography{custom}

\clearpage
\appendix
\section{Prompt for Data Format Conversion}
\label{sec:ap_prompt}
In this section, we list the prompts used during the data conversion process, where the entire data format transformation is performed using the GPT-4o API. We employed two types of conversions: (1) converting the original annotated data into a multiple-choice question format, and (2) transforming yes/no type questions into a bidirectional format.

\begin{tcolorbox}[
    colback=blue!10,
    colframe=blue!50!black,
    title=Prompt for multiple-choice questions,
    fonttitle=\bfseries,
    drop shadow=black!50!white,
    sharp corners,
    boxrule=0.8mm,
    breakable,
    label={box:prompt_1}
]

\textbf{[Task]}: The task is to modify a question to make it significantly harder and more nuanced. The revised question should focus on specific technical aspects, avoid straightforward clues, and include plausible distractors that are incorrect. Use the correct answer as a subtle hint in the question, while making the other options appear equally viable.\\

\textbf{[Guideline]}: \\Original question: ``\texttt{\{original\_question\}}" \\
Correct answer: ``\texttt{\{correct\_answer\}}" \\
Options: ``\texttt{\{options\}}" \\

\textbf{[Requirement]}: \\Please rewrite this question by following these strict guidelines:
\begin{itemize}
    \item Focus on a specific, less obvious aspect that indirectly hints at the correct answer.
    \item Add realistic but irrelevant details to make the question look more complex.
    \item Avoid directly mirroring the wording of the correct answer, making the question more ambiguous.
    \item Ensure the distractor options contain minor technical errors or realistic details that seem logical but are incorrect.
    \item Add only contextual details related to the original question and correct answer; avoid introducing unrelated information.
    \item Do not change the correct answer.
\end{itemize}
\textbf{[Output format]}:
Your response must be a single line, formatted exactly as follows: \\
``\texttt{...(modified question), Choices: A) xxx B) xxx C) xxx D) xxx}" \\
Only this format is allowed, and any deviations from this format are strictly forbidden.

\end{tcolorbox}

\begin{tcolorbox}[
    colback=blue!10,
    colframe=blue!50!black,
    title=Prompt for yes/no type questions,
    fonttitle=\bfseries,
    drop shadow=black!50!white,
    sharp corners,
    boxrule=0.8mm,
    breakable,
    label={box:prompt_2}
]

\textbf{[Task]}: Strictly follow the instructions. Rewrite the following question by reversing the temporal sequence and format it as a multiple-choice question with the following choices: A. Yes B. No.\\

\textbf{[Guideline]}: \\Original question: ``\texttt{\{original\_question\}}" \\

\textbf{[Requirement]}: 
\begin{itemize}
    \item Reverse the temporal sequence of the original question while keeping it grammatically correct.
    \item Format the rewritten question as a multiple-choice question with the choices A. Yes and B. No.
    \item Ensure the output strictly follows the specified format without deviations.
\end{itemize}

\textbf{[Output format]}: 
Your response must be a single line, formatted exactly as follows: \\
``\texttt{...(statement).Choices: A.Yes B.No}" \\

Only this format is allowed, and any deviations from this format are strictly forbidden.

\end{tcolorbox}

\section{Dataset Case}
\label{sec:ap_case}
\begin{figure*}[h]
  \centering
  \includegraphics[width=0.95\textwidth]{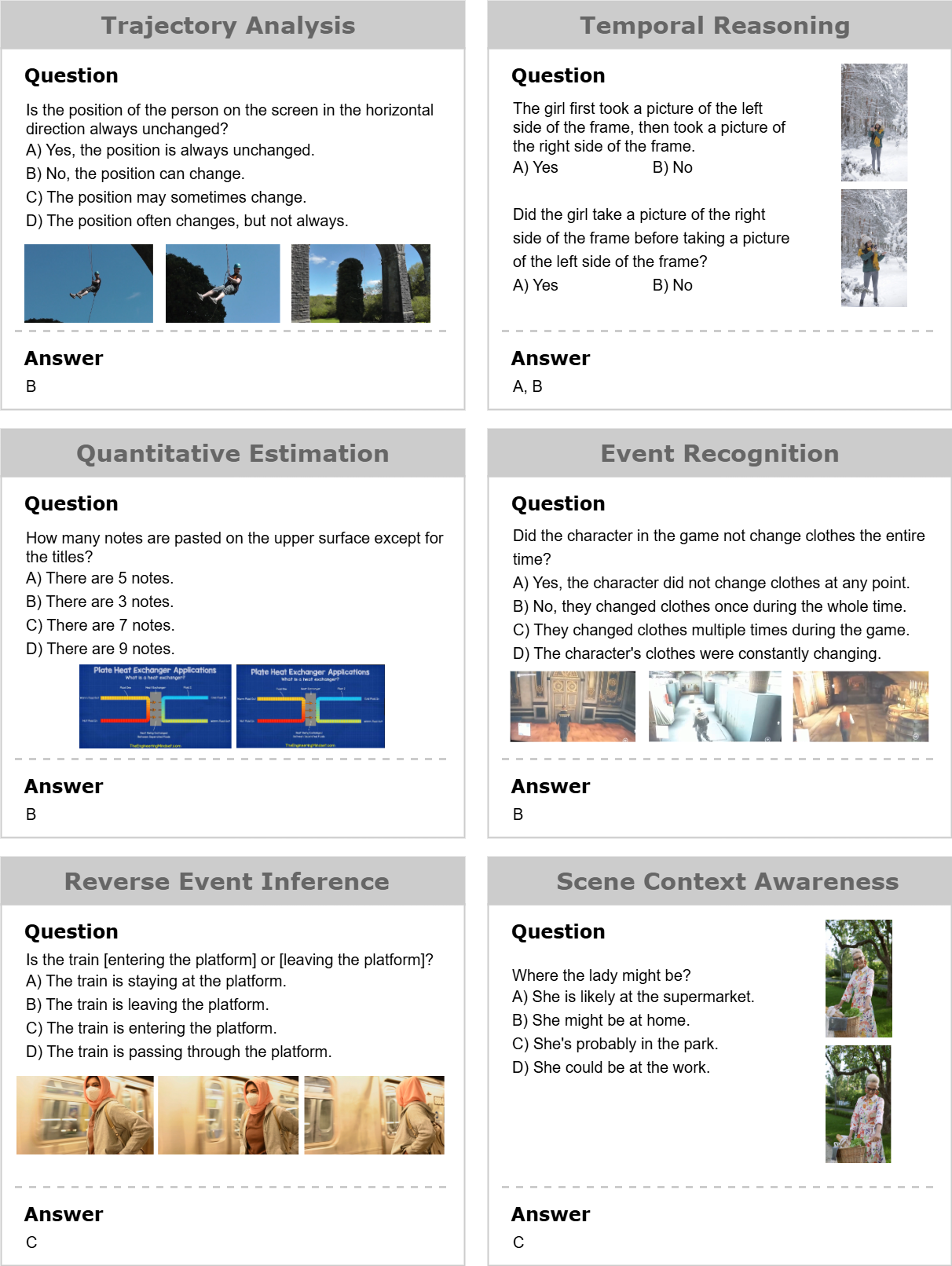}
\caption{Dataset cases 1.}
  \label{ex_case_1}
\end{figure*}

\begin{figure*}[h]
  \centering
  \includegraphics[width=0.95\textwidth]{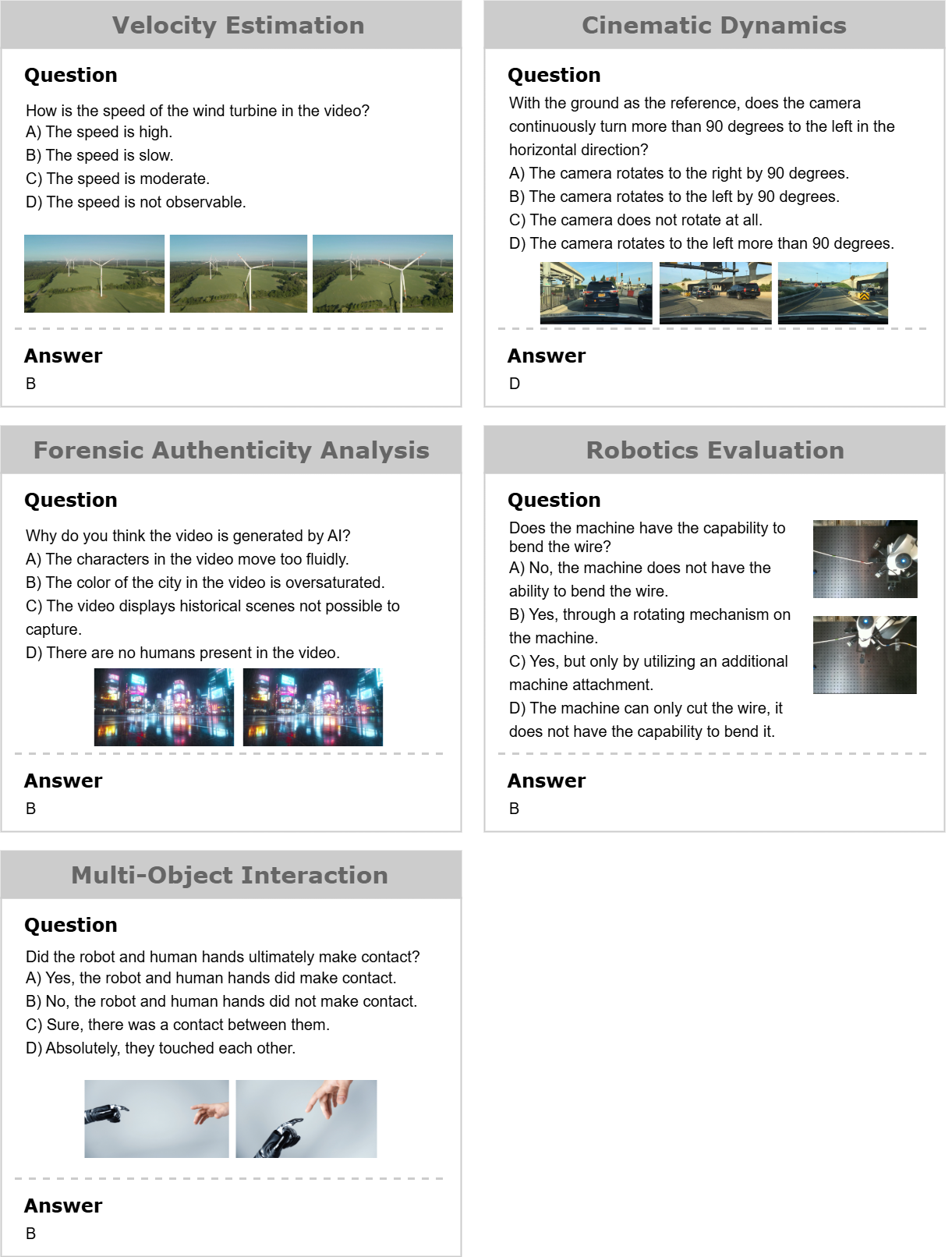}
\caption{Dataset cases 2.}
  \label{ex_case_2}
\end{figure*}

In this subsection, we present the constructed examples for each category in \ours, as shown in Figure \ref{ex_case_1} and Figure \ref{ex_case_2}.

\end{document}